\pdfoutput=1
\documentclass[10pt,twocolumn,letterpaper]{article}

\usepackage{cvpr}
\usepackage{times}
\usepackage{epsfig}
\usepackage{graphicx}
\usepackage{amsmath}
\usepackage{amssymb}

 \usepackage{color}
\usepackage[pagebackref=true,breaklinks=true,letterpaper=true,colorlinks,bookmarks=false]{hyperref}

\cvprfinalcopy 

\ifcvprfinal\fi

\begin{document}

\title{Deepfakes Detection with Automatic Face Weighting}


\author{Daniel Mas Montserrat, Hanxiang Hao,  S. K. Yarlagadda, Sriram Baireddy, Ruiting Shao \\ J{\'a}nos Horv{\'a}th, Emily Bartusiak, Justin Yang, David G{\"u}era, Fengqing Zhu, Edward J. Delp\\  \\
Video and Image Processing Laboratory (VIPER)\\
School of Electrical Engineering\\
Purdue University\\
West Lafayette, Indiana, USA\\
}

\maketitle
\thispagestyle{empty}

\begin{abstract}
Altered and manipulated 
 multimedia is increasingly present and widely distributed via social media platforms. 
 Advanced video manipulation tools enable the generation of highly realistic-looking altered multimedia. 
While many methods have been presented to detect manipulations, most of them fail when evaluated with data outside of the datasets used in research environments. 
In order to address this problem, the Deepfake Detection Challenge (DFDC) provides a large dataset of videos containing realistic manipulations and an evaluation system that ensures that methods work quickly and accurately, even when faced with challenging data.
In this paper, we introduce a method based on convolutional neural networks (CNNs) and recurrent neural networks (RNNs) that extracts visual and temporal features from faces present in videos to accurately detect manipulations. 
The method is evaluated with the DFDC dataset, providing competitive results compared to other techniques.

\end{abstract}

\section{Introduction}
Manipulated multimedia is rapidly increasing its presence on the Internet and social media. Its rise is fueled by the mass availability of easy-to-use tools and techniques for generating realistic fake multimedia content. 
Recent advancements in the field of deep learning have led to the development of methods to create artificial images and videos that are eerily similar to authentic images and videos. Manipulated multimedia created using such techniques typically involving neural networks, such as Generative Adversarial Networks (GAN)~\cite{goodfellow2014generative} and Auto-Encoders (AE)~\cite{Goodfellow-et-al-2016}, are generally referred to as Deepfakes. 
While these tools can be useful to automate steps in movie production, video game design, or virtual reality rendering, they are potentially very damaging if used for malicious purposes.  
As manipulation tools become more accessible, realistic, and undetectable, the divide between real and fake multimedia is blurred. Furthermore, social media allows for the uncontrolled spread of manipulated content at a large scale. This spread of misinformation damages journalism and news providers as it gets increasingly difficult to distinguish between reliable and untrustworthy information sources. 

\begin{figure}[t]
	\centering
	\includegraphics[width=0.99\columnwidth]{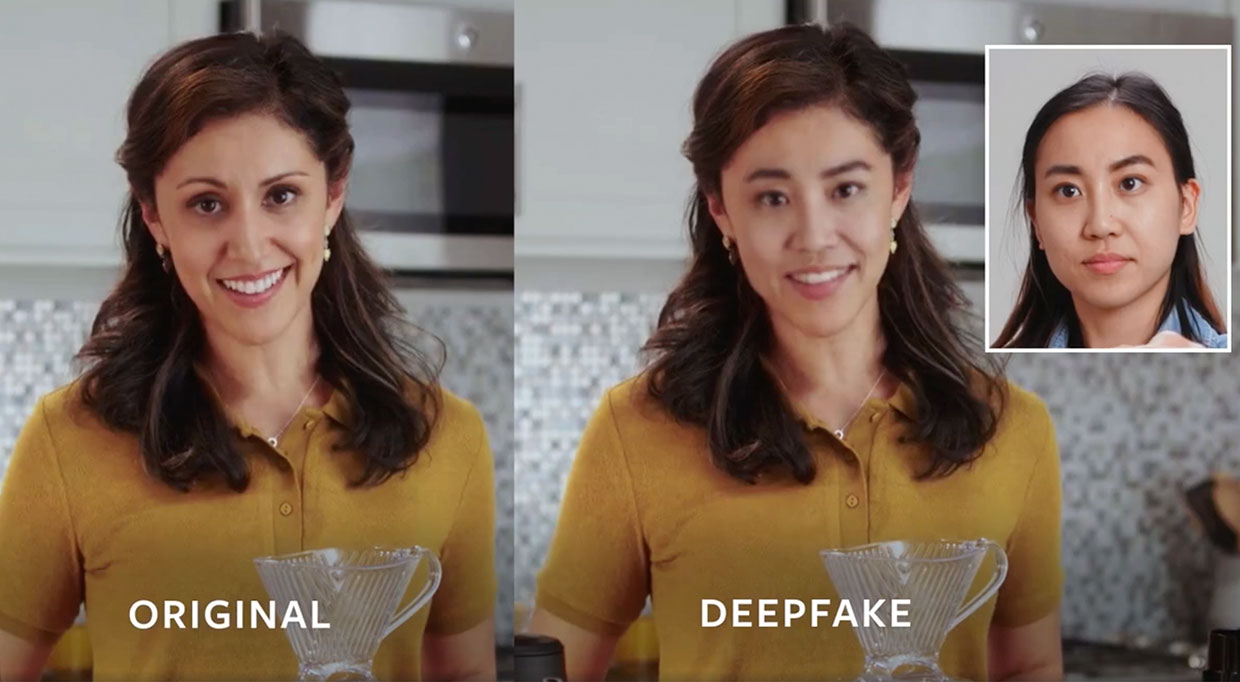} 
	\label{fig:large_forg_int}
	\caption{Example of images from DFDC~\cite{dolhansky2019deepfake} dataset: original image (left) and manipulated image with the swapped face (right).}
\end{figure}

Human facial manipulations are among the most common Deepfake forgeries. Through face swaps, an individual can be placed at some location he or she was never present at. By altering the lip movement and the associated speech signal, realistic videos can be generated of individuals saying words they actually never uttered. This type of Deepfake manipulation can be very damaging when used to generate graphic adult content or fake news that can alter the public opinion. In fact, many images and videos containing such Deepfake forgeries are already present on adult content web sites, news articles, and social media.

Image and video manipulations have been utilized for a long time. Before the advent of Deepfakes, editing tools such as Photoshop \cite{photoshop_2016} or GIMP \cite{gimp} have been widely used for image manipulations. Some common forgeries include splicing (inserting objects into images)~\cite{cozzolino2019noiseprint}, copy and moving parts within an image (copy-move forgery)~\cite{barni2019copy}, or shadow removal~\cite{yarlagadda2019shadow}. While research on detecting such manipulations has been conducted for more than a decade \cite{cozzolino2019noiseprint, cozzolino2015splicebuster, yarlagadda2018satellite, yarlagadda2019shadow, bartusiak2019splicing, horvath2019anomaly, barni2017aligned, barni2019copy, Guera2019_ICMLW}, many techniques fail to detect more recent and realistic manipulations, especially when the multimedia alterations are performed with deep learning methods. 
Fortunately, there is an increasing effort to develop reliable detection technology 
such as AWS, Facebook, Microsoft, and the Partnership on AI’s Media Integrity Steering Committee with the Deepfake Detection Challenge (DFDC)~\cite{dolhansky2019deepfake}.

Advances in deep learning have resulted in a great variety of methods that have provided groundbreaking results in many areas including computer vision, natural language processing, and biomedical applications \cite{nature}. While several neural networks that detect a wide range of manipulations have been introduced \cite{marra2018detection, marra2019gans, zhang2019detecting, yu2019attributing, afchar2018mesonet, rossler2019faceforensics++}, new generative methods that create very realistic fake multimedia \cite{hui2020image, le2019shadow, brock2018large, deepfakes, neuraltextures} are presented every year, leading to a push and pull problem where manipulation methods try to fool new detection methods and vice-versa. Therefore, there is a need for methods that are capable of detecting multimedia manipulations in a robust and rapid manner.

In this paper, we present a novel model architecture that combines a Convolutional Neural Network (CNN) with a Recurrent Neural Network (RNN) to accurately detect facial manipulations in videos. The network automatically selects the most reliable frames to detect these manipulations with a weighting mechanism combined with a Gated Recurrent Unit (GRU) that provides a final probability of a video being real or being fake. We train and evaluate our method with the Deepfake Detection Challenge dataset, obtaining a final score of 0.321 (log-likelihood error, the lower the better) at position 117 of 2275 teams (top 6\%) of the public leader-board.

\section{Related Work}

There are many techniques for face manipulation and generation. Some of the most commonly used include FaceSwap~\cite{faceswap}, Face2Face~\cite{face2face}, DeepFakes~\cite{deepfakes}, and NeuralTextures~\cite{neuraltextures}. FaceSwap and Face2Face are computer graphics based methods while the other two are learning based methods. In FaceSwap \cite{faceswap}, a face from a source video is projected onto a face in a target video using facial landmark information. The face is successfully projected by minimizing the difference between the projected shape and the target face's landmarks. Finally, the rendered face is color corrected and blended with the target video. In Face2Face~\cite{face2face}, facial expressions from a selected face in a source video are transferred to a face in the target video. Face2Face uses selected frames from each video to create dense reconstructions of the two faces. These dense reconstructions are used to re-synthesize the target face with different expressions under different lighting conditions. In DeepFakes~\cite{deepfakes}, two autoencoders~\cite{Goodfellow-et-al-2016} (with a shared encoder) are trained to reconstruct target and source faces. To create fake faces, the trained encoder and decoder of the source face are applied on the target face. This fake face is blended onto the target video using Poisson image editing~\cite{perez2003poisson}, creating a Deepfake video. Note the difference between DeepFakes (capital F), the technique now being described, and Deepfakes (lowercase f), which is a general term for fake media generated with deep learning-based methods. In NeuralTextures~\cite{neuraltextures}, a neural texture of the face in the target video is learned. This information is used to render the facial expressions from the source video on the target video.

In recent years, methods have been developed to detect such deep learning-based manipulations. In~\cite{marra2018detection}, several CNN architectures have been tested in a supervised setting to discriminate between GAN generated images and real images. Preliminary results are promising but the performance degrades as the difference between training and testing increases or when the data is compressed. In ~\cite{marra2019gans, zhang2019detecting, yu2019attributing}, forensic analysis of GAN generated images revealed that GANs leave some high frequency fingerprints in the images they generate.

Additionally, several techniques to detect videos containing facial manipulations have been presented. While some of these methods focus on detecting videos containing only DeepFake manipulations, others are designed to be agnostic to the technique used to perform the facial manipulation. The work presented in~\cite{guera2018deepfake, sabir2019recurrent} use a temporal-aware pipeline composed by a Convolutional Neural Network (CNN) and a Recurrent Neural Network (RNN) to detect DeepFake videos. 
Current DeepFake videos are created by splicing synthesized face regions onto the original video frames. This splicing operation can leave artifacts that can later be detected when estimating the 3D head pose. The authors of~\cite{yang2019exposing} exploit this fact and use the difference between the head pose estimated with the full set of facial landmarks and a subset of them to separate DeepFake videos from real videos. This method provided competitive results on the UADFV~\cite{li2018ictu} database.
The same authors proposed a method~\cite{li2018exposing} to detect DeepFake videos by analyzing the face warping artifacts.
The authors of~\cite{afchar2018mesonet} detect manipulated videos generated by the DeepFake and Face2Face techniques with a shallow neural network that acts on mesoscopic features extracted from the video frames to distinguish manipulated videos from real ones. However, the results presented in~\cite{rossler2019faceforensics++} demonstrated that in a supervised setting, several deep network based models~\cite{huang2017densely, szegedy2016rethinking, chollet2017xception} outperform the ones based on shallow networks when detecting fake videos generated with DeepFake, Face2Face, FaceSwap, and NeuralTexture.

\begin{figure*}[t]
	\centering
	\includegraphics[width=1.0\textwidth]{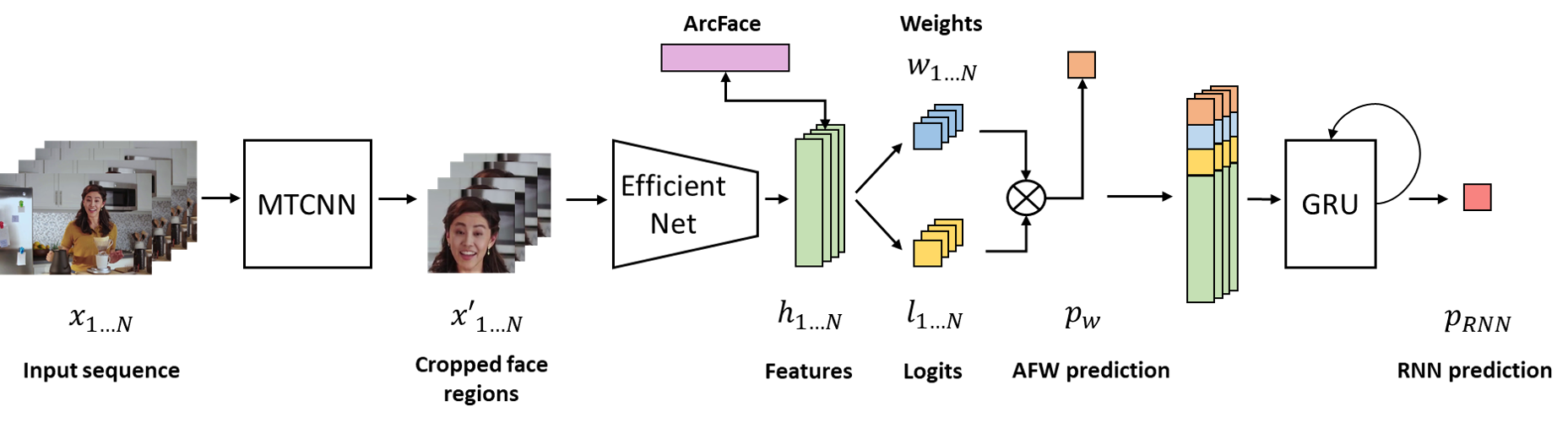}
	\caption{Block Diagram of our proposed Deepfake detection system: MTCNN detects faces within the input frames, then EfficientNet extracts features from all the detected face regions, and finally the Automatic Face Weighting (AFW) layer and the Gated Recurrent Unit (GRU) predict if the video is real or manipulated.}
	\label{fig:method_sum}
\end{figure*}

\section{Deepfake Detection Challenge Dataset}
\label{sec:dataset}

The Deepfake Detection Challenge (DFDC)~\cite{dolhansky2019deepfake} dataset contains a total of 123,546 videos with face and audio manipulations. Each video contains one or more people and has a length of 10 seconds with a total of 300 frames. The nature of these videos typically includes standing or sitting people, either facing the camera or not, with a wide range of backgrounds, illumination conditions, and video quality. The training videos have a resolution of $1920 \times 1080$ pixels, or $1080 \times 1920$ pixels if recorded in vertical mode. Figure \ref{fig:large_forg_int} shows some examples of frames from videos of the dataset. This dataset is composed by a total of 119,146 videos with a unique label (real or fake) in a training set, 400 videos on the validation set without labels and 4000 private videos in a testing set. The 4000 videos of the test set can not be inspected but models can be evaluated on it through the Kaggle system. The ratio of manipulated:real videos is 1:0.28. Because only the 119,245 training videos contain labels, we use the totality of that dataset to train and validate our method. The provided training videos are divided into 50 numbered parts. We use 30 parts for training, 10 for validation and 10 for testing.

A unique label is assigned to each video specifying whether it contains a manipulation or not. However, it is not specified which type of manipulation is performed: face, audio, or both. As our method only uses video information, manipulated videos with only audio manipulations will lead to noisy labels as the video will be labeled as fake but faces will be real. Furthermore, more than one person might be present in the video, with face manipulations performed on only one of them.

The private set used for testing evaluates submitted methods within the Kaggle system and reports a log-likelihood loss. Log-likelihood loss drastically penalizes being both confident and wrong. In the worst case, a prediction that a video is authentic when it is actually manipulated, or the other way around, will add infinity to your error score. In practice, if this worst-case happens, the loss is clipped to a very big value. This evaluation system poses an extra challenge, as methods with good performance in metrics like accuracy, could have very high log-likelihood errors.

\section{Proposed Method}

Our proposed method (Figure \ref{fig:method_sum}) extracts visual and temporal features from faces by using a combination of a CNN with an RNN. Because all visual manipulations are located within face regions, and faces are typically present in a small region of the frame, using a network that extracts features from the entire frame is not ideal. Instead, we focus on extracting features only in regions where a face is present. Because networks trained with general image classification task datasets such as ImageNet \cite{imagenet} have performed well when transferred to other tasks \cite{huh2016makes}, we use pre-trained backbone networks as our starting point. Such backbone networks extract features from faces that are later fed to an RNN to extract temporal information. The method has three distinct steps: (1) face detection across multiple frames using MTCNN~\cite{Zhang_2016}, (2) feature extraction with a CNN, and (3) prediction estimation with a layer we refer to as Automatic Face Weighting (AFW) along with a Gated Recurrent Unit (GRU). 
Our approach is described in detail in the following subsections, including a boosting and test augmentation approach we included in our DFDC submission.

\subsection{Face Detection}

We use MTCNN~\cite{Zhang_2016} to perform face detection.
MTCNN is a multi-task cascaded model that can produce both face bounding boxes and facial landmarks simultaneously.
The model uses a cascaded three-stage architecture to predict face and landmark locations in a coarse-to-fine manner. 
Initially, an image pyramid is generated by resizing the input image to different scales.
The first stage of MTCNN then obtains the initial candidates of facial bounding boxes and landmarks given the input image pyramid. 
The second stage takes the initial candidates from the first stage as the input and rejects a large number of false alarms.
The third stage is similar to the second stage but with a larger input image size and deeper structure to obtain the final bounding boxes and landmark points. 
Non-maximum suppression and bounding box regression are used in all three stages to remove highly overlapped candidates and refine the prediction results.
With the cascaded structure, MTCNN refines the results stage by stage in order to get accurate predictions.

We choose this model because it provides good detection performance on both real and synthetic faces in the DFDC dataset. 
While we also considered more recent methods like BlazeFace~\cite{Bazarevsky_2019}, which provides faster inferencing, its false positive rate on the DFDC dataset was considerably larger than that of MTCNN.

We extract faces from 1 every 10 frames for each video. In order to speed up the face detection process, we downscale the frame by a factor of 4. 
Additionally, we include a margin of 20 pixels at each side of the detected bounding boxes in order to capture a broader area of the head as some regions such as the hair might contain artifacts useful to detect manipulations.
After processing the input frames with MTCNN, we crop all the regions where faces were detected and resize them to 224 $\times$ 224 pixels.

\subsection{Face Feature Extraction}

After detecting face regions, a binary classification model is trained to extract features that can be used to classify the real/fake faces.
The large number of videos that have to be processed in a finite amount of time for the Deepfake Detection Challenge requires networks that are both fast and accurate. In this work, we use EfficientNet-b5 \cite{efficientnet} as it provides a good trade-off between network parameters and classification accuracy. Additionally, the network has been designed using neural architecture search (NAS) algorithms, resulting in a network that is both compact and accurate. In fact, this network has outperformed previous state-of-the-art approaches in datasets such as ImageNet \cite{imagenet} while having fewer parameters.

Since the DFDC dataset contains many high-quality photo-realistic fake faces, discriminating between real and manipulated faces can be challenging. 
To achieve a better and more robust face feature extraction, we combine EfficientNet with the additive angular margin loss (also known as ArcFace)~\cite{Deng_2019} instead of a regular softmax+cross-entropy loss. ArcFace is a learnable loss function that is based on the classification cross-entropy loss but includes penalization terms to provide a more compact representation of the categories.
ArcFace simultaneously reduces the intra-class difference and enlarges the inter-class difference between the classification features.
It is designed to enforce a margin between the distance of the sample to its class center and the distances of the sample to the centers of other classes in an angular space.
Therefore, by minimizing the ArcFace loss, the classification model can obtain highly discriminative features for real faces and fake faces to achieve a more robust classification that succeeds even for high-quality fake faces.

\subsection{Automatic Face Weighting}
While an image classification CNN provides a prediction for a single image, we need to assign a prediction for an entire video, not just a single frame. The natural choice is to average the predictions across all frames to obtain a video-level prediction. However, this approach has several drawbacks. First, face detectors such as MTCNN can erroneously report that background regions of the frames contain faces, providing false positives. Second, some videos might include more than one face but with only one of them being manipulated. Furthermore, some frames might contain blurry faces where the presence of manipulations might be difficult to detect. In such scenarios, a CNN could provide a correct prediction for each frame but an incorrect video-level prediction after averaging.

In order to address this problem, we propose an automatic weighting mechanism to emphasize the most reliable regions where faces have been detected and discard the least reliable ones when determining a video-level prediction. This approach, similar to attention mechanisms \cite{vaswani2017attention}, automatically assigns a weight, $w_j$, to each logit, $l_j$, outputted by the EfficientNet network for each $j$th face region. Then, these weights are used to perform a weighted average of all logits, from all face regions found in all sampled frames to obtain a final probability value of the video being fake. Both logits and weights are estimated using a fully-connected linear layer with the features extracted by EfficientNet as input. In other words, the features extracted by EfficientNet are used to estimate a logit (that indicates if the face is real or fake) and a weight (that can provide information of how confident or reliable is the logit prediction). The output probability, $p_w$, of a video being false, by the automatic face weighting is:

\begin{equation}
\label{eq:face-weighting}
p_w = \sigma( \frac{\sum_{j=1}^{N} w_j l_j}{\sum_{j=1}^{N} w_j} )
\end{equation}

Where $w_j$ and $l_j$ are the weight value and logit obtained for the $j$th face region, respectively and $\sigma(.)$ is the Sigmoid function. Note that after the fully-connected layer, $w_j$ is passed through a ReLU activation function to enforce that $w_j \geq 0$. Additionally, a very small value is added to avoid divisions by 0. This weighted sum aggregates all the estimated logits providing a video-level prediction.

\subsection{Gated Recurrent Unit}
The backbone model estimates a logit and weight for each frame without using information from other frames. While the automatic face weighting combines the estimates of multiple frames, these estimates are obtained by using single-frame information. However, ideally the video-level prediction would be performed using information from all sampled frames.

In order to merge the features from all face regions and frames, we include a Recurrent Neural Network (RNN) on top of the automatic face weighting. We use a Gated Recurrent Unit (GRU) to combine the features, logits, and weights of all face regions to obtain a final estimate.
For each face region, the GRU takes as input a vector of dimension 2051 consisting of the features extracted from EfficientNet (with dimension 2048), the estimated logit $l_j$, the estimated weighting value $w_j$, and the estimated manipulated probability after the automatic face weighting $p_w$. 
Although $l_j$, $w_j$, $p_w$, and the feature vectors are correlated, we input all of them to the GRU and let the network itself extract the useful information.
The GRU is composed of 3 stacked bi-directional layers and a uni-directional layer with a hidden layer with dimension 512. The output of the last layer of the GRU is mapped through a linear layer and a Sigmoid function to estimate a final probability $p_{RNN}$ of the video being manipulated.

\subsection{Training Process}
We use a pre-trained MTCNN for face detection and we only train our EfficientNet, GRU, and the Automatic Face Weighting layers. The EfficientNet is initialized with weights pre-trained on ImageNet. The GRU and AFW layers are initialized with random weights. During the training process, we oversample real videos (containing only unmanipulated faces) to balance the dataset. The network is trained end-to-end with 3 distinct loss functions: an ArcFace loss with the output of EfficentNet, a binary cross-entropy loss with the automatic face weighting prediction $p_w$, and a binary cross-entropy loss with the GRU prediction $p_{RNN}$. 

The ArcFace loss is used to train the EfficientNet layers with batches of cropped faces from randomly selected frames and videos. This loss allows the network to learn from a large variety of manipulated and original faces with various colors, poses, and illumination conditions. Note that ArcFace only trains the layers from EfficientNet and not the GRU layers or the fully-connected layers that output the AFW weight values and logits.

The binary cross-entropy (BCE) loss is applied at the outputs of the automatic face weighting layer and the GRU. The BCE loss is computed with cropped faces from frames of a randomly selected video. Note that this loss is based on the output probabilities of videos being manipulated (video-level prediction), while ArcFace is a loss based on frame-level predictions. The BCE applied to $p_w$ updates the EfficientNet and AFW weights. The BCE applied to $p_{RNN}$ updates all weights of the ensemble (excluding MTCNN). 

While we train the complete ensemble end-to-end, we start the training process with an optional initial step consisting of 2000 batches of random crops applied to the ArcFace loss to obtain an initial set of parameters of the EfficientNet. This initial step provides the network with useful layers to later train the automatic face weighting layer and the GRU. While this did not present any increase in detection accuracy during our experiments, it provided a faster convergence and a more stable training process.

Due to computing limitations of GPUs, the size of the network, and the number of input frames, only one video can be processed at a time during training. However, the network parameters are updated after processing every 64 videos (for the binary cross-entropy losses) and 256 random frames (for the ArcFace loss). We use Adam as the optimization technique with a learning rate of 0.001.

\begin{figure*}[ht]
	\centering
	\includegraphics[width=1.0\textwidth]{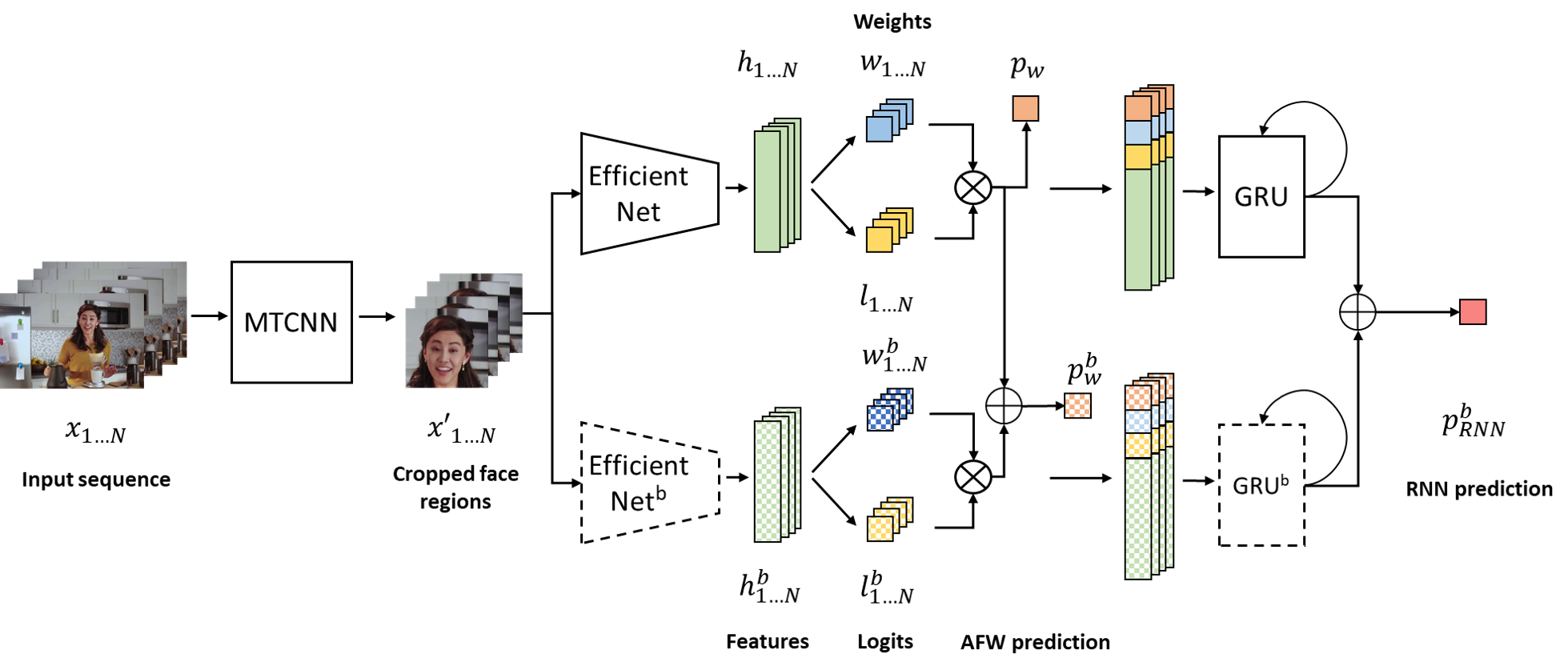}
	\caption{Diagram of the proposed method including the boosting network (dashed elements). The predictions of the main and boosting network are combined at the AFW layer and after the GRUs. We train the main network with the training set and the boosting network with the validation set. }
	\label{fig:boosting}
\end{figure*}

\subsection{Boosting Network}
The logarithmic nature of the binary cross-entropy loss (or log-likelihood error) used at the DFDC leads to large penalizations for predictions both confident and incorrect. In order to obtain a small log-likelihood error we want a method that has both good detection accuracy and is not overconfident of its predictions. In order to do so, we use two main approaches during testing: (1) adding a boosting network and (2) applying data augmentation during testing.

The boosting network is a replica of the previously described network. However, this auxiliary network is not trained to minimize the binary cross-entropy of the real/fake classification, but trained to predict the error between the predictions of our main network and the ground truth labels. We do so by estimating the error of the main network on the logit domain for both the AFW and GRU outputs. When using the boosting network, the prediction outputted by the automatic face weighting layer, $p^b_w$, is defined as:

\begin{equation}
\label{eq:boosting-1}
p^b_w = \sigma( \frac{\sum_{j=1}^{N} (w_j l_j + w^b_j l^b_j)}{\sum_{j=1}^{N} (w_j + w^{b}_j)} )
\end{equation}

Where $w_j$ and $l_j$ are the weights and logits outputted by the main network and  $w^b_j$ and $l^b_j$, are the weights and logits outputted by the boosting network for the $j$th input face region and $\sigma(.)$ is the Sigmoid function. In a similar manner, the prediction outputted by the GRU, $p^{b}_{RNN}$, is:

\begin{equation}
\label{eq:boosting-2}
p^b_{RNN} = \sigma( l_{RNN} + l^b_{RNN} )
\end{equation}

Where $l_{RNN}$ is the logit outputted by the GRU of the main network, $l^b_{RNN}$ is the logit outputted by the GRU of the boosting network, and $\sigma(.)$ is the Sigmoid function.

While the main network is trained using the training split of the dataset, described in section \ref{sec:dataset}, we train the boosting network with the validation split.

Figure \ref{fig:boosting} presents the complete diagram of our system when including the boosting network. The dashed elements and the symbols with superscripts form part of the boosting network. The main network and the boosting network are combined at two different points: at the automatic face weighting layer, as described in equation \ref{eq:boosting-1}, and after the gated recurrent units, as described in equation \ref{eq:boosting-2}.


\subsection{Test Time Augmentation}
Besides adding the boosting network, we perform data augmentation during testing. For each face region detected by the MTCNN, we crop the same region in the 2 previous and 2 following frames of the frame being analyzed. Therefore we have a total of 5 sequences of detected face regions. We run the network within each of the 5 sequences and perform a horizontal flip in some of the sequences randomly. Then, we average the prediction of all the sequences. This approach helps to smooth out overconfident predictions: if the predictions of different sequences disagree, averaging all the probabilities leads to a lower number of both incorrect and overconfident predictions.

\section{Experimental Results}
We train and evaluate our method with the DFDC dataset, described in section \ref{sec:dataset}. Additionally, we compare the presented approach with 4 other techniques. We compare it with the work presented in \cite{guera2018deepfake} and a modified version that only process face regions detected by MTCNN. We also evaluate two CNNs: EfficientNet \cite{efficientnet} and Xception \cite{chollet2017xception}. For these networks, we simply average the predictions for each frame to obtain a video-level prediction. 

We use the validation set to select the configuration for each models that provides the best balanced accuracy. Table \ref{table:accuracy} presents the results of balanced accuracy. Because it is based on extracting features on the entire video, Conv-LSTM \cite{guera2018deepfake} is unable to capture the manipulations that happen within face regions. However, if the method is adapted to process only face regions, the detection accuracy improves considerably. Classification networks such as Xception \cite{chollet2017xception}, which provided state-of-the-art results in FaceForensics++ dataset \cite{rossler2019faceforensics++}, and EfficientNet-b5 \cite{efficientnet} show good accuracy results. Our work shows that by including an automatic face weighting layer and a GRU, the accuracy is further improved.

\begin{table}[htp]
  \centering
    \caption{Balanced accuracy of the presented method and previous works.}
    \label{table:accuracy}
    \vspace{0.1cm}
  \begin{tabular}{l c c}
        \textbf{Method}     & \textbf{Validation}    &  \textbf{Test} \\
    \hline 
    \textbf{Conv-LSTM} \cite{guera2018deepfake}       & 55.82\% & 57.63\%     \\
    \textbf{Conv-LSTM} \cite{guera2018deepfake} \textbf{+ MTCNN}      & 66.05\% & 70.78\%     \\
    \textbf{EfficientNet-b5} \cite{efficientnet}     & 79.25\% & 80.62\%      \\
    \textbf{Xception} \cite{chollet2017xception}    & 78.42\% & 80.14\%      \\
    \textbf{Ours}     & 92.61\% & 91.88\%      \\
  \end{tabular}
\end{table}

Additionally, we evaluate the accuracy of the predictions at every stage of our method. Table \ref{table:accuracy-2} shows the balanced accuracy of the prediction obtained by the averaging the logits predicted by EfficientNet, $l_j$ (logits), the prediction of the automatic face weighting layer, $p_w$ (AFW), and the prediction after the gated recurrent unit, $p_{RNN}$ (GRU). We can observe that every stage increases the detection accuracy, obtaining the highest accuracy with the GRU prediction.

\begin{table}[htp]
  \centering
    \caption{Balanced accuracy of at different stages of our method.}
    \label{table:accuracy-2}
    \vspace{0.1cm}
  \begin{tabular}{l c c}
        \textbf{Method}     & \textbf{Validation Accuracy}\\
    \hline 
    \textbf{Ours (logits)}     & 85.51\%    \\
    \textbf{Ours (AFW)}     & 87.90\% \\
    \textbf{Ours (GRU)}     & 92.61\% \\
  \end{tabular}

\end{table}

\begin{figure}[t]
	\centering
	\includegraphics[width=1.0\columnwidth]{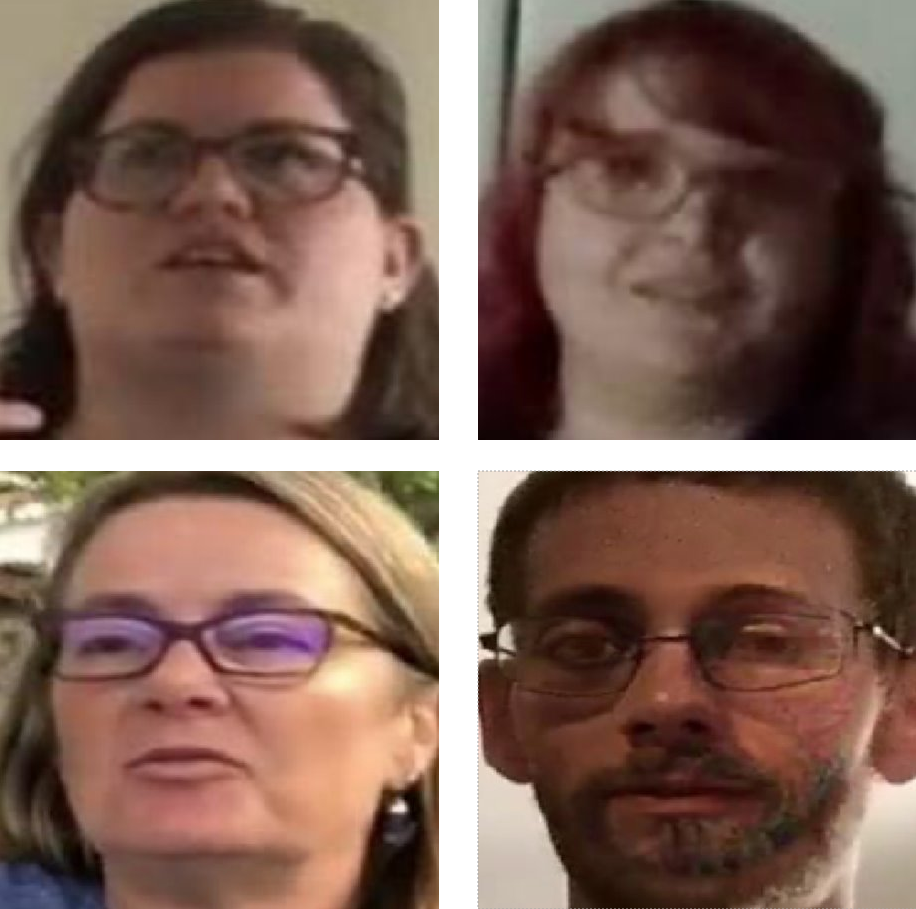}
	\caption{Examples of faces with manipulations from DFDC. The images in the top are incorrectly classified by the network. The bottom images are correctly classified.}
	\label{fig:results}
\end{figure}

Figure \ref{fig:results} shows some examples of correctly (bottom) and incorrectly (top) detected manipulations. We observed that the network typically fails when faced with highly-realistic manipulations that are performed in blurry or low-quality images. Manipulations performed in high-quality videos seem to be properly detected, even the challenging ones.

We evaluate the effect of using the boosting network and data augmentation during testing. In order to so, we use the private testing set on the Kaggle system and report our log-likelihood error (the lower the better). Table \ref{table:log} shows that by using both the boosting and test augmentation we are able to decrease our log-likelihood down to 0.321. This place the method in the position 117 of 2275 teams (5.1\%) of the competition's public leader-board.

\begin{table}[htp]
  
  \centering
     \caption{The log-likelihood error of our method with and without boosting network and test augmentation.}
     \label{table:log}
     \vspace{0.1cm}
  \begin{tabular}{l c}
        \textbf{Method}     & \textbf{Log-likelihood}\\
    \hline 
    \textbf{Baseline}      & 0.364 \     \\
    \textbf{+ Boosting Network}     & 0.341     \\
    \textbf{+ Test Augmentation}     & 0.321      \\

  \end{tabular}

\end{table}

\section{Conclusions}
In this paper, we present a new method to detect face manipulations within videos. 
We show that combining convolutional and recurrent neural networks achieves high detection accuracies on the DFDC dataset. 
We describe a method to automatically weight different face regions and boosting techniques can be used to obtain more robust predictions. 
The method processes videos quickly (in less than eight seconds) with a single GPU. 

Although the results of our experiments are promising, new techniques to generate deepfake manipulations emerge continuously. 
The modular nature of the proposed approach allows for many improvements, such as using different face detection methods, different backbone architectures, and other techniques to obtain a prediction from features of multiple frames. 
Furthermore, this work focuses on face manipulation detection and dismisses any analysis of audio content which could provide a significant improvement of detection accuracy in future work.

\section{Acknowledgment}
This material is based on research sponsored by DARPA
and Air Force Research Laboratory (AFRL) under agreement number FA8750-16-2-0173. 
The U.S. Government is authorized to reproduce and distribute reprints for Governmental purposes notwithstanding any copyright notation thereon. 
The views and conclusions contained herein are those of the authors and should not be interpreted as necessarily representing the official policies or endorsements, either expressed or implied, of DARPA and Air Force Research Laboratory (AFRL) or the U.S. Government.

Address all correspondence to Edward J. Delp, ace@ecn.purdue.edu .

{\small
\bibliographystyle{IEEEbib}
\bibliography{ref}
}

\end{document}